\documentclass[a4paper]{article}

\makeatother

\usepackage{amsmath}

\usepackage{graphicx}
\usepackage{amsfonts}
\usepackage{amssymb}
\usepackage{url}
\usepackage{bm}
\usepackage{float}
\usepackage{algorithm}
\usepackage{algpseudocode}
\usepackage{multirow}
\usepackage{textcomp}
\usepackage{bbm}
\usepackage[table,xcdraw]{xcolor}

\title{Machine Speech Chain with One-shot Speaker Adaptation}

\author{Andros Tjandra$^{1,2}$, Sakriani Sakti$^{1,2}$, Satoshi Nakamura$^{1,2}$\\
	$^1$Nara Institute of Science and Technology\\Graduate School of Information Science, Japan\\
	$^2$RIKEN, Center for Advanced Intelligence Project AIP, Japan \\
	\texttt{\{andros.tjandra.ai6,ssakti,s-nakamura\}@is.naist.jp}
}

\begin{document}

\maketitle
\begin{abstract}

In previous work, we developed a closed-loop speech chain model based on deep learning, in which the architecture enabled the automatic speech recognition (ASR) and text-to-speech synthesis (TTS) components to mutually improve their performance. This was accomplished by the two parts teaching each other using both labeled and unlabeled data. This approach could significantly improve model performance within a single-speaker speech dataset, but only a slight increase could be gained in multi-speaker tasks. Furthermore, the model is still unable to handle unseen speakers. In this paper, we present a new speech chain mechanism by integrating a speaker recognition model inside the loop. We also propose extending the capability of TTS to handle unseen speakers by implementing one-shot speaker adaptation. This enables TTS to mimic voice characteristics from one speaker to another with only a one-shot speaker sample, even from a text without any speaker information. In the speech chain loop mechanism, ASR also benefits from the ability to further learn an arbitrary speaker’s characteristics from the generated speech waveform, resulting in a significant improvement in the recognition rate.
\end{abstract}

\section{Introduction}

In human communication, a closed-loop speech chain mechanism has a critical auditory feedback mechanism from the speaker's mouth to her ear \cite{denes1993speech}. In other words, the hearing process is critical not only for the listener but also for the speaker. By simultaneously listening and speaking, the speaker can monitor the volume, articulation, and general comprehensibility of her speech. Inspired by such a mechanism, we previously constructed a machine speech chain \cite{tjandra2017speechchain} based on deep learning. This architecture enabled ASR and TTS to mutually improve their performance by teaching each other.

One of the advantages of using a machine speech chain is the ability to train a model based on the concatenation of both labeled and unlabeled data. For supervised training with labeled data (speech-text pair data), both ASR and TTS models can be trained independently by minimizing the loss of their predicted target sequence and the ground truth sequence. However, for unsupervised training with unlabeled or unpaired data (speech only or text only), the two models need to support each other through a connection. Our experimental results reveal that such a connection enabled ASR and TTS to further improve their performance by using only unpaired data. Although this technique could provide a significant improvement in model performance within a single-speaker speech dataset, only a slight increase could be gained in multi-speaker tasks.

Difficulties arise due to the fundamental differences in the ASR and TTS mechanisms. The ASR task is to ``extract''  data from a large amount of information and only retain the spoken content (many-to-one mapping). On the other hand, the TTS task aims to ``generate'' data from compact text information into a generated speech waveform with an arbitrary speaker’s characteristics and speaking style (one-to-many mapping). The imbalanced amounts of information contained inside the text and speech causes information loss inside the speech-chain and hinders us in perfectly reconstructing the original speech. To enable the TTS system to mimic the voices of different speakers, we previously only added speaker information via a speaker's identity by one-hot encoding. However, this is not a practical solution because we are still unable to handle unseen speakers.

In this paper, we propose a new approach to handle voice characteristics from an unknown speaker and minimize the information loss between speech and text inside the speech chain loop. First, we integrate a speaker recognition system into the speech chain loop. Second, we extend the capability of TTS to handle the unseen speaker using one-shot speaker adaptation. This enables TTS to mimic voice characteristics from one speaker to another with only a one-shot speaker sample, even from text without any speaker information. In the speech chain loop mechanism, ASR also benefits from furthering learning an arbitrary speaker’s characteristics from the generated speech waveform. We evaluated our proposed model with the well-known Wall Street Journal corpus, consisting of multi-speaker speech utterances that are often used as an ASR benchmark test set. Our new speech mechanism is able to handle unseen speakers and improve the performance of both the ASR and TTS models.

\section{Machine Speech Chain Framework} \label{sec:speechchain_deepspk}

\begin{figure*}[]

	\centering
	\includegraphics[width=1.0\linewidth]{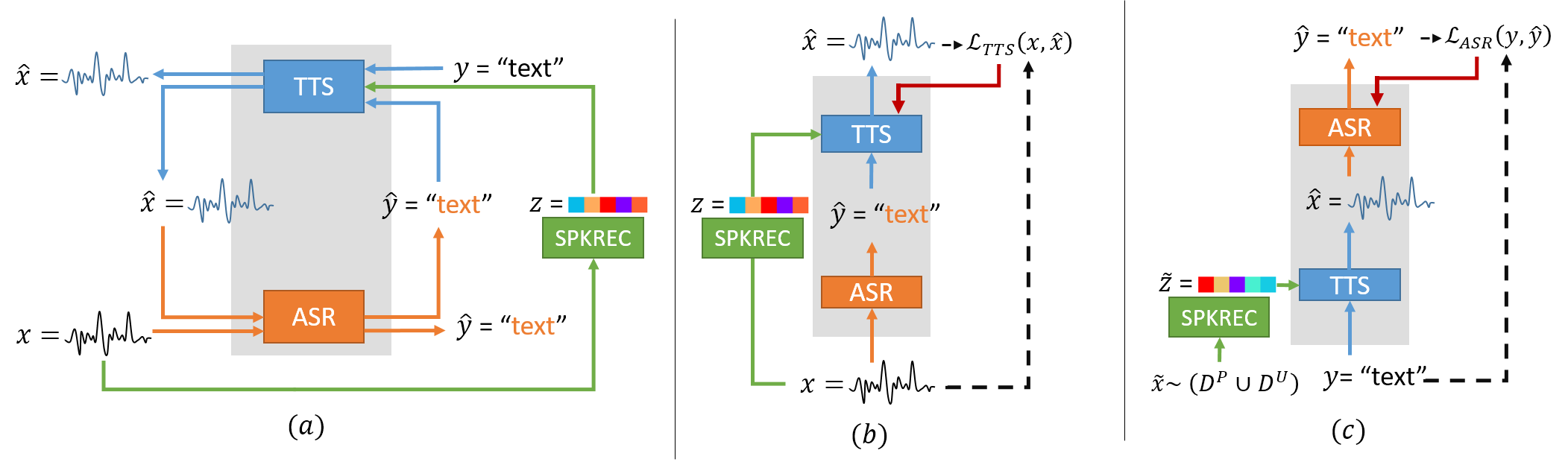}
	\caption{Overview of proposed machine speech chain architecture with speaker recognition; (b) Unrolled process with only speech utterances and no text transcription (speech $\rightarrow$ \textbf{[ASR,SPKREC]} $\rightarrow$ [text + speaker vector] $\rightarrow$ \textbf{TTS} $\rightarrow$ speech); (c) Unrolled process with only text but no corresponding speech utterance ([text + speaker vector by sampling \textbf{SPKREC}] $\rightarrow$ \textbf{TTS} $\rightarrow$ speech $\rightarrow$ \textbf{ASR} $\rightarrow$ text). Note: grayed box is the original speech chain mechanism.}
	
	\label{fig:speech_chain_with_spkrec}
	
\end{figure*}
Figure.~\ref{fig:speech_chain_with_spkrec} illustrates the new speech chain mechanism. Similar to the earlier version, it consists of a sequence-to-sequence ASR \cite{bahdanau2016end, chan2016listen}, a sequence-to-sequence TTS \cite{wang2017tacotron}, and a loop connection from ASR to TTS and from TTS to ASR. The key idea is to jointly train the ASR and TTS models. The difference is that in this new version, we integrate a speaker recognition model inside the loop illustrated in Fig.~\ref{fig:speech_chain_with_spkrec}(a). As mentioned above, we can train our model on the concatenation of both labeled (paired) and unlabeled (unpaired) data. In the following, we describe the learning process.
\begin{enumerate}
	
	\item \textbf{Paired speech-text dataset (see Fig.~\ref{fig:speech_chain_with_spkrec}(a))}
	Given the speech utterances $\mathbf{x}$ and the corresponding text transcription $\mathbf{y}$ from dataset $\mathcal{D}^P$, both ASR and TTS models can be trained independently. Here, we can train ASR by calculating the ASR loss $L_{ASR}^P$ directly with teacher-forcing. For TTS training, we generate a speaker embedding vector $z = \text{SPKREC}(\mathbf{x})$, integrate $z$ information with the TTS, and calculate the TTS loss $L_{TTS}^P$ via teacher-forcing.
	
	\item \textbf{Unpaired speech data only  (see Fig.~\ref{fig:speech_chain_with_spkrec}(b))}
	Given only the speech utterances $\mathbf{x}$ from unpaired dataset $\mathcal{D}^U$, ASR generates the text transcription $\mathbf{\hat{y}}$ (with greedy or beam-search decoding), and SPKREC provides a speaker-embedding vector $z = \text{SPKREC}(\mathbf{x})$. TTS then reconstructs the speech waveform $\mathbf{\hat{x}} = TTS(\mathbf{\hat{y}}, z)$, and given the generated text $\mathbf{\hat{y}}$ and the original speaker vector $z$ via teacher forcing. After that, we calculate the loss $L_{TTS}^U$ between $\mathbf{x}$ and $\mathbf{\hat{x}}$.
	
	\item \textbf{Unpaired text data only (see Fig.~\ref{fig:speech_chain_with_spkrec}(c))}
	Given only the text transcription $\mathbf{y}$ from unpaired dataset $\mathcal{D}^U$, we need to sample speech from the available dataset $\mathbf{\tilde{x}} \sim (\mathcal{D}^P \cup \mathcal{D}^U)$ and generate a speaker vector $\tilde{z} = \text{SPKREC}(\tilde{x})$ from SPKREC. Then, the TTS generates the speech utterance $\mathbf{\hat{x}}$ with greedy decoding, while the ASR reconstructs the text $\mathbf{\hat{y}} = ASR(\mathbf{\hat{x}})$, given generated speech $\mathbf{\hat{x}}$ via teacher forcing. After that, we calculate the loss $L_{ASR}^U$ between $\mathbf{y}$ and $\mathbf{\hat{y}}$.
	
\end{enumerate}
We combine all loss together and update both ASR and TTS model:

\begin{align}
L = \alpha * (L^P_{ASR} &+ L^P_{TTS}) + \beta * (L^U_{ASR} + L^U_{TTS}) \\
\theta_{ASR} &= Optim(\theta_{ASR}, \nabla_{\theta_{ASR}}L)\\
\theta_{TTS} &= Optim(\theta_{TTS}, \nabla_{\theta_{TTS}}L)
\end{align} where $\alpha, \beta$ are hyper-parameters to scale the loss between supervised (paired) and unsupervised (unpaired) loss. 

\section{Sequence-to-Sequence ASR}

A sequence-to-sequence \cite{sutskever2014sequence} architecture is a type of neural network that directly models the conditional probability $P(\mathbf{y}|\mathbf{x})$ between two sequences $\mathbf{x}$ and $\mathbf{y}$. For an ASR model, we assume the source sequence $\mathbf{x} = [x_1,..,x_S]$ is a sequence of speech feature (e.g., Mel-spectrogram, MFCC) and the target sequence $\mathbf{y} = [y_1,..,y_T]$ is a sequence of grapheme or phoneme.

The encoder reads source speech sequence $\mathbf{x}$, forwards it through several layers (e.g., LSTM\cite{hochreiter1997long}/GRU\cite{chung2014empirical}, convolution), and extracts high-level feature representation $\mathbf{h}^e=[h_1^e,..,h_S^e]$ for the decoder. The decoder is an autoregressive model that produces the current output conditioned on the previous output and the encoder states $\mathbf{h}^e$. To bridge the information between decoder states $h_t^d$ and encoder states $\mathbf{h}^e$, we use an attention mechanism \cite{bahdanau2014neural} to calculate the alignment probability $a_t(s) = Align(h_s^e, h_t^d); \enskip \forall s\in[1..S]$ and then calculate the expected context vector $c_t = \sum_{s=1}^{S} a_t(s) * h_s^e$. Finally, the decoder predicts the target sequence probability $p_{y_t} = P(y_t|c_t, h_t^d, \mathbf{y}_{<t}; \theta_{ASR})$. In the training stage, we optimized the ASR by minimizing the negative log-likelihood loss function:

\begin{equation}
\mathcal{L}_{ASR}(y, p_{y}) = -\sum_{t=1}^{T}\sum_{c=1}^{C}\mathbbm{1}(y_t=c)*\log{(p_{y_t}[c])}
\end{equation}

\section{Sequence-to-Sequence TTS with One-shot Speaker Adaptation} \label{sec:tts_new} 

A parametric TTS can be formulated as a sequence-to-sequence model where the source sequence is a text utterance $\mathbf{y}=[y_1,..,y_T]$ with length $T$, and the target sequence is a speech feature $\mathbf{x} = [x_1,..,x_S]$ with length $S$. Our model objective is to maximize $P(\mathbf{x}|\mathbf{y}; \theta_{TTS})$ w.r.t TTS parameter $\theta_{TTS}$. We build our model upon the basic structure of the ``Tacotron" TTS \cite{wang2017tacotron} and ``DeepSpeaker" \cite{li2017deep} models.

The original Tacotron is a single speaker TTS system based on a sequence-to-sequence model. Given a text utterance, Tacotron produces the Mel-spectrogram and the linear spectrogram followed by the Griffin-Lim algorithm to recover the phase and reconstruct the speech signal. However, the original model is not designed to incorporate speaker identity or to generate speech from different speakers.

On the other hand, DeepSpeaker is a deep neural speaker-embedding system (here denoted as ``SPKREC"). Given a sequence of speech features $\mathbf{x} = [x_1, ..,x_S]$, DeepSpeaker generates an L2-normalized continuous vector embedding $z$. If $\mathbf{x}_1$ and $\mathbf{x}_2$ are spoken by the same speakers, the trained DeepSpeaker model will produce the vector $z_1 = \text{SPKREC}(\mathbf{x}_1)$ and the vector $z_2 = \text{SPKREC}(\mathbf{x}_2)$, which are close to each other. Otherwise, the generated embeddings $z_1$ and $z_2$ will be far from each other. By combining Tacotron with DeepSpeaker, we can do {``one-shot''} speaker adaptation by conditioning the Tacotron with the generated fixed-size continuous vector $z$ from the DeepSpeaker with a single speech utterance from any speaker.

\begin{figure}[]
	\centering
	\includegraphics[width=0.9\linewidth]{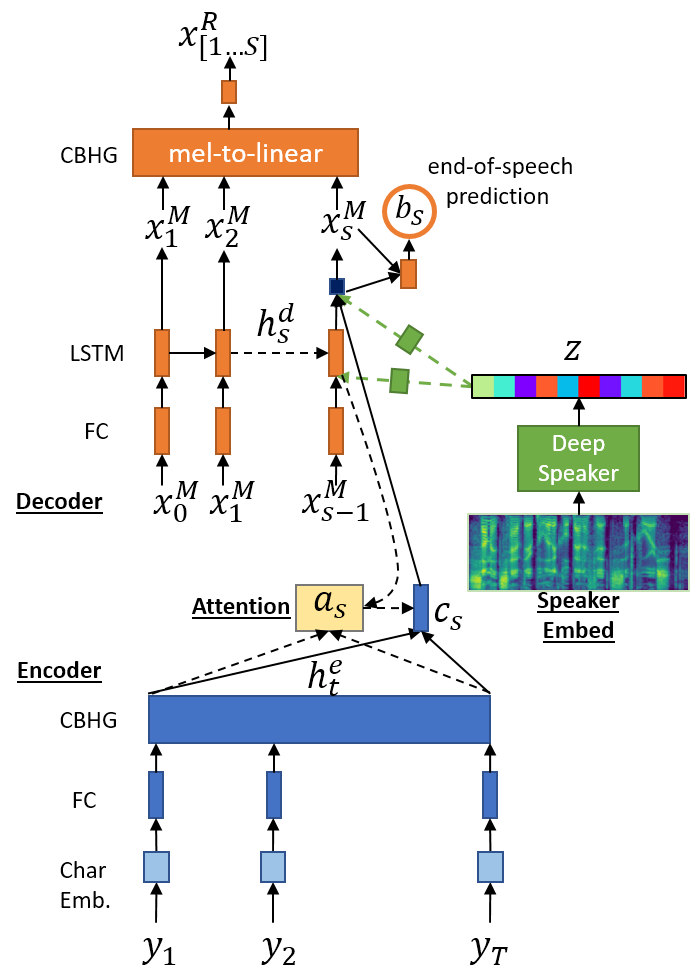}
	
	\caption{Proposed model: sequence-to-sequence TTS (Tacotron) + speaker information via neural speaker embedding (DeepSpeaker).}
	\label{fig:tacotron_deepspk}
	
\end{figure}

Here, we adopt both systems by modifying the original Tacotron TTS model to integrate the DeepSpeaker model. Figure~\ref{fig:tacotron_deepspk} illustrates our proposed model. From the encoder module, the character embedding maps a sequence of characters into a continuous vector. The continuous vector is then projected by two fully connected (FC) layers with the LReLU\cite{xu2015empirical} function. We pass the results to a CBHG module (1D \textbf{C}onvolution \textbf{B}ank + \textbf{H}ighway + bidirectional \textbf{G}RU) with $K$=8 (1 to 8) different filter sizes. The final output $\mathbf{h}^e = [h^e_1,..h^e_T]$ from the CBHG module represents high-level information from input text $\mathbf{y}$.

On the decoder side, we have an autoregressive decoder that produces current output Mel-spectrogram $\hat{x}_s^M$ given the previous output $x_{s-1}^M$, the encoder context vector $c_{t}$, and the speaker-embedding vector $z$. First, at the time-step $s$-th, the previous input $x_{s-1}^M$ is projected by two FC layers with LReLU. Then, to inform our decoder which speaker style will be produced, we feed the corresponding speech utterance and generate speaker-embedding vector $z = SPKREC(\mathbf{x}^M)$. This speaker embedding $z$ is generated by using only 1 utterance of target speakers, thus it is called as ``{one-shot}'' speaker adaptation. After that, we integrate the speaker vector $z$ with a linear projection and sum it with the last output from the FC layer. Then, we apply two LSTM layers to generate current decoder state $h^d_s$. To retrieve the relevant information between the current decoder state and the entire encoder state, we calculate the attention probability $a_s(t) = Align(h^e_t, h^d_s); \forall t\in[1..T]$ and the expected context vector $c_s = \sum_{1}^{T} a_s(t) * h_t^e$. Then, we concatenate the decoder state $h^d_s$, context vector $c_s$, and projected speaker-embedding $z$ together into a vector, followed by two fully connected layers to produce the current time-step Mel-spectrogram output $x_s^M$. Finally, all predicted outputs of Mel-spectrogram $\mathbf{x}^M =[x_1^M, .., x_S^M]$ are projected into a CBHG module to invert the corresponding Mel-spectrogram into a linear-spectrogram $\mathbf{x}^R=[x_1^R, .., x_S^R]$. Additionally, we have an end-of-speech prediction module to predict when the speech is finished. The end-of-speech prediction module reads the predicted Mel-spectrogram $\hat{x}_s^M$ and the context vector $c_s$, followed by an FC layer and sigmoid function to produce a scalar $b_s \in [0..1]$.

In the training stage, we optimized our proposed model by minimizing the following loss function: 

\begin{align}
\mathcal{L}_{TTS}(.) =& \left( \sum_{s=1}^{S} \gamma_{1} \left( \Vert x_s^M - \hat{x}_s^M \Vert^2_2 + \Vert x_s^R - \hat{x}_s^R \Vert^2_2 \right) \right. \nonumber \\
& \left. - \gamma_{2} \left  (b_s \log(\hat{b}_s) + (1-b_s) \log(1-\hat{b}_s)\right) \right) \nonumber \\
&+ \gamma_{3} \left(1 -\frac{\left< \hat{z}, z \right> }{\Vert \hat{z} \Vert_2 \enskip \Vert z \Vert_2}\right) \label{eq:loss_tts_1}
\end{align} where $\gamma_{1}, \gamma_{2}, \gamma_{3}$ are our sub-loss hyper-parameters, and $\mathbf{x}^M, \mathbf{x}^R, {b}, {z}$ are the ground truth Mel-spectrogram, linear spectrogram, and end-of-speech label and speaker-embedding vector from real speech data, respectively. $\mathbf{\hat{x}}^M, \mathbf{\hat{x}}^R, \hat{b}$ represent the predicted Mel-spectrogram, linear spectrogram, and end-of-speech label, respectively, and speaker-embedding vector $\hat{z} = \text{SPKREC}(\hat{\mathbf{x}}^M)$ is the predicted speaker vector from the Tacotron output. Here, $\mathcal{L}_{\theta_{TTS}}$ consists of 3 different loss formulations: Eq.~\ref{eq:loss_tts_1} line 1 applies L2 norm-squared error between ground-truth and predicted speech as a regression task, Eq.~\ref{eq:loss_tts_1} line 2 applies binary cross entropy for end-of-speech prediction as a classification task, and Eq.~\ref{eq:loss_tts_1} line 3 applies cosine distance between the ground-truth speaker-embedding $z$ and predicted speaker-embedding $\hat{z}$, which is the common metric for measuring the similarity between two vectors; furthermore, by minimizing this loss, we also minimize the global loss of speaker style \cite{larsen2016autoencoding, johnson2016perceptual}.

\section{Experiment}

\subsection{Corpus Dataset}

In this study, we run our experiment on the Wall Street Journal CSR Corpus \cite{paul1992design}. The complete data are contained in an SI284 (SI84+SI200) dataset. We follow the standard Kaldi \cite{povey11asru} s5 recipe to split the training set, development set, and test set. To reformulate the speech chain as a semi-supervised learning method, we prepare SI84 and SI200 as paired and unpaired training sets, respectively. SI84 consists of 7138 utterances (about 16 hours of speech) spoken by 83 speakers, and SI200 consists of 30,180 utterances (about 66 hours) spoken by 200 speakers (without any overlap with speakers of SI84). We use ``dev93" to denote the development and ``eval92" for the test set.

\subsection{Feature and Text Representation}

All raw speech waveforms are represented at a 16-kHz sampling rate. We extracted two different sets of features. First, we applied pre-emphasis (0.97) on the raw waveform, and then we extracted the log-linear spectrogram with 50-ms window length, 12.5-ms step size, and 2048-point short-time Fourier transform (STFT) with the Librosa package \cite{librosa}. Second, we extracted the log Mel-spectrogram with an 80 Mel-scale filterbank. For our TTS model, we used both log-linear and log-Mel spectrogram for the first and second output. For our ASR and DeepSpeaker model, we used the log-Mel spectrogram for the encoder input.

The text utterances were tokenized as characters and mapped into a 33-character set: 26 alphabetic letters (a-z), 3 punctuation marks ('.-), and 4 special tags $\langle$\texttt{noise}$\rangle$, $\langle$\texttt{spc}$\rangle$,$\langle$\texttt{s}$\rangle$, and $\langle$\texttt{/s}$\rangle$ as noise, space, start-of-sequence, and end-of-sequence tokens, respectively. Both ASR input and TTS output shared the same text representation.

\subsection{Model Details}

For the ASR model, we used a standard sequence-to-sequence model with an attention module. On the encoder sides, the input log Mel-spectrogram features were processed by 3 bidirectional LSTMs (Bi-LSTM) with 256 hidden units for each LSTM (total 512 hidden units for Bi-LSTM). To reduce memory consumption and processing time, we used hierarchical sub-sampling \cite{graves2012supervised, bahdanau2016end} on all three Bi-LSTM layers and thus reduced the sequence length by a factor of 8. On the decoder sides, we projected the one-hot encoding from the previous character into a 256-dims continuous vector with an embedding matrix, followed by one unidirectional LSTM with 512 hidden units. For the attention module, we used standard content-based attention \cite{bahdanau2014neural}. In the decoding phase, the transcription was generated by beam-search decoding (size=5), and we normalized the log-likelihood score by dividing it with its own length to prevent the decoder from favoring the shorter transcriptions. We did not use any language model or lexicon dictionary in this work.

For the TTS model, we used the proposed TTS explained in Sec.~\ref{sec:tts_new}. The hyperparameters for the basic structure were generally the same as those for the original Tacotron, except we replaced ReLU with the LReLU function. For the CBHG module, we used $K=8$ filter banks instead of 16 to reduce the GPU memory consumption. For the decoder sides, we deployed two LSTMs instead of GRU with 256 hidden units. For each time-step, our model generated 4 consecutive frames to reduce the number of steps in the decoding process. For the sub-loss hyperparameter in Eq.~\ref{eq:loss_tts_1}, we set $\gamma_{1}=1,\gamma_{2}=1,\gamma_{3}=0.25$.

For the speaker recognition model, we used the DeepSpeaker model and followed the original hyper-parameters in the previous paper. However, our DeepSpeaker is only trained on the WSJ SI84 set with 83 unique speakers. Thus, the model is expected to generalize effectively across all remaining unseen speakers to assist the TTS and speech chain training. We used the Adam optimizer with a learning rate of $5e-4$ for the ASR and TTS models and $1e-3$ for the DeepSpeaker model. All of our models in this paper are implemented with PyTorch \cite{paszke2017automatic}.

\section{Experiment Result}

\begin{table}[]
	\centering
	\caption{Character error rate (CER (\%)) comparison between results of supervised learning and those of a semi-supervised learning method, evaluated on \textit{test\_eval92} set}
	\label{tbl:asr}
	
	\begin{tabular}{|l|c|}
		\hline
		\multicolumn{1}{|c|}{\textbf{Model}}                                              & \textbf{CER (\%)}                                        \\ \hline \hline
\multicolumn{2}{|c|}{\cellcolor[HTML]{EFEFEF}\textbf{\begin{tabular}[c]{@{}c@{}}Supervised training: \\WSJ \textit{train\_si84} (paired) $\rightarrow$ Baseline\end{tabular}}}                  \\ \hline 

		Att Enc-Dec \cite{kim2017joint}                                                 & 17.01                                                    \\ \hline
		Att Enc-Dec \cite{tjandra2017wav2text}                                          & 17.68                                                    \\ \hline
		Att Enc-Dec (ours)                                                                & 17.35                                                    \\ \hline\hline
\multicolumn{2}{|c|}{\cellcolor[HTML]{EFEFEF}\textbf{\begin{tabular}[c]{@{}c@{}}Supervised training: \\WSJ \textit{train\_si284} (paired) $\rightarrow$ Upperbound\end{tabular}}} \\ \hline

		Att Enc-Dec \cite{kim2017joint}                                                 & 8.17                                                     \\ \hline
		Att Enc-Dec \cite{tjandra2017wav2text}                                          & 7.69                                                     \\ \hline
		Att Enc-Dec (ours)                                                                & 7.12                                                     \\ \hline\hline
\multicolumn{2}{|c|}{\cellcolor[HTML]{EFEFEF}\textbf{\begin{tabular}[c]{@{}c@{}}Semi-supervised training: \\WSJ \textit{train\_si84} (paired) + \textit{train\_si200} (unpaired)\end{tabular}}}          \\ \hline

		Label propagation (greedy)                                                         & 17.52                                                    \\ \hline
		Label propagation (beam=5)                                                         & 14.58                                                    \\ \hline
		\textbf{Proposed speech chain (Sec.~\ref{sec:speechchain_deepspk}) }                                           & \textbf{9.86}                                                     \\ \hline
	\end{tabular}

\end{table}
\begin{table}[]
	\centering
	\caption{L2-norm squared on log-Mel spectrogram to compare the supervised learning and those of a semi-supervised learning method, evaluated on \textit{test\_eval92} set. {Note: We did not include standard Tacotron (without SPKREC) into the table since it could not output various target speaker.}}
	\label{tbl:tts}

	\begin{tabular}{|l|c|}
		\hline
		\multicolumn{1}{|c|}{\textbf{Model}}                                              & \textbf{L2-norm$^2$}                                        \\ \hline \hline
		\multicolumn{2}{|c|}{\cellcolor[HTML]{EFEFEF}\textbf{\begin{tabular}[c]{@{}c@{}}Supervised training: \\WSJ \textit{train\_si84} (paired) $\rightarrow$ Baseline\end{tabular}}}                  \\ \hline 
		Proposed Tacotron (Sec.~\ref{sec:tts_new}) (ours)                                                                & 1.036                                                    \\ \hline\hline
		\multicolumn{2}{|c|}{\cellcolor[HTML]{EFEFEF}\textbf{\begin{tabular}[c]{@{}c@{}}Supervised training: \\WSJ \textit{train\_si284} (paired) $\rightarrow$ Upperbound\end{tabular}}} \\ \hline
		Proposed Tacotron (Sec.~\ref{sec:tts_new}) (ours)                                                               & 0.836                                                     \\ \hline\hline
		\multicolumn{2}{|c|}{\cellcolor[HTML]{EFEFEF}\textbf{\begin{tabular}[c]{@{}c@{}}Semi-supervised training: \\WSJ \textit{train\_si84} (paired) + \textit{train\_si200} (unpaired)\end{tabular}}}          \\ \hline
		\textbf{Proposed speech chain (Sec.~\ref{sec:speechchain_deepspk} + Sec.~\ref{sec:tts_new}) }                                           & \textbf{0.886}                                                     \\ \hline
	\end{tabular}
	
\end{table}

Table~\ref{tbl:asr} shows the ASR results from multiple scenarios evaluated on eval92. In the first block, we trained our baseline model by using paired samples from the SI84 set only, and we achieved 17.35\% CER. In the second block, we trained our model with paired data of the full WSJ SI284 data, and we achieved 7.12\% CER as our upper-bound performance. In the last block, we trained our model with a semi-supervised learning approach using SI84 as paired data and SI200 as unpaired data. For comparison with other models trained with semi-supervised learning, we carried out a label-propagation \cite{zhu02learningfrom} strategy to ``generate'' the ground-truth for the unlabeled speech dataset, and the model with beam-size=5 successfully reduced the CER to 14.58\%. Nevertheless, our proposed speech-chain model could achieve a significant improvement over all baselines (paired only and label-propagation) with 9.86\% CER, close to the upper-bound results.

Table~\ref{tbl:tts} shows the TTS results from multiple scenarios evaluated on eval92. We calculate the difference with L2-norm squared between ground-truth and and the predicted log-Mel spectrogram. We observed similar trends with the ASR results, where the semi-supervised training with speech chain method improved significantly over the baseline, and close to the upper-bound result.

\section{Related Works}

While single speaker TTS has achieved high-quality results \cite{wang2017tacotron, oord2016wavenet}, speaker adaptation remained a challenging task for TTS system. As discussed in \cite{swietojanski2014learning}, adaptation techniques for neural networks fall into three classes: feature-space transformation, auxiliary features augmentation, and model-based adaptation. Wu et al. \cite{wu2015study}
performed a systematic speaker adaptation for DNN-based speech synthesis
at different levels. First, i-vector features \cite{dehak2011front} to represent speaker identity was augmented at the input level. Then, they performed model adaptation using the learning hidden unit contributions at the middle level based on the speaker dependent parameters \cite{swietojanski2014learning}. Finally, feature space transformations are applied at the output level. The parameters are transformed to mimic the target speaker’s voice with joint density Gaussian mixture model (JD-GMM) model \cite{toda2007voice}.

Our adaptation approach might fall into a similar category with the 
augmentation of auxiliary features such an i-vector. But, in this case, we utilize DeepSpeaker \cite{li2017deep} that is trained to minimize the distance between embedding pairs from the same speaker and maximize the distance between pairs from different speakers. It has been proved to provide better performance on speaker recognition task compare to i-vector. Furthermore, instead of focusing a speaker adaptation task only on TTS, we integrate all end-to-end models including ASR, TTS, and DeepSpeaker into a machine speech chain loop.

\section{Conclusion}

In this paper, we introduce a new speech chain mechanism by integrating a speaker recognition model inside the loop. By using the new proposed system, we eliminate the downside from our previous speech chain, where we are unable to incorporate the data from unseen speakers. We also extending the capability of TTS to generate speech from unseen speaker by implementing the one-shot speaker adaptation. Thus, the TTS can generate a speech with a similar voice characteristic only with a one-shot speaker example. Inside the speech chain loop, the ASR also get new data from the combination between a text sentence and an arbitrary voice characteristic. Our results shows that after we deploy the speech-chain loop, the ASR system got significant improvement compared to the baseline (supervised training only) and other semi-supervised technique (label propagation). Similar trends as ASR, the TTS system also got an improvement compared to the baseline (supervised training only).

\section{Acknowledgment}

Part of this work was supported by JSPS KAKENHI Grant Numbers JP17H06101 and JP17K00237.
\bibliographystyle{IEEEtran}
\bibliography{refs}

\end{document}